\documentclass{new_tlp}

\usepackage{graphicx}
\usepackage{mathptmx}
\usepackage{amsmath}
\usepackage{amssymb}
\usepackage{url}
\usepackage{multicol}
\usepackage{framed}
\usepackage{float}

\usepackage{tikz}
\usetikzlibrary{positioning,shapes,shadows,arrows,calc}

\tikzstyle{activity}=[rectangle, draw=black, rounded corners, text centered, text width=8em, fill=white, drop shadow]
\tikzstyle{data}=[rectangle, draw=black, text centered, fill=black!10, text width=8em, drop shadow]
\tikzstyle{myarrow}=[->, thick]
\pgfdeclarelayer{background}
\pgfdeclarelayer{foreground}
\pgfsetlayers{background,main,foreground}

\newcommand{\clasp}{\texttt{clasp}}
\newcommand{\claspre}{\texttt{claspre}}
\newcommand{\claspresta}{\texttt{claspre(s)}}
\newcommand{\claspredyn}{\texttt{claspre(s+d)}}
\newcommand{\gringo}{\texttt{gringo}}

\newcommand{\aspeed}{\texttt{aspeed}}
\newcommand{\claspfolio}[1]{\texttt{claspfolio}}
\newcommand{\claspfolioTwo}[1]{\texttt{claspfolio}~2}

\newcommand{\satzilla}{\texttt{SATzilla}}
\newcommand{\hydra}{\texttt{Hydra}}
\newcommand{\isac}{\texttt{ISAC}}
\newcommand{\sss}{\texttt{3S}}
\newcommand{\measp}{\texttt{ME-ASP}}
\newcommand{\cphydra}{\texttt{CP-Hydra}}

\newcommand{\stonesoup}{\texttt{Stone Soup}}
\newcommand{\snapp}{\texttt{SNAPP}}
\newcommand{\llama}{\texttt{LLAMA}}
\newcommand{\aqme}{\texttt{AQME}}

\newcommand{\claspfoliolike}[1]{\texttt{claspfolio-#1-like}}
\newcommand{\satzillalike}[1]{\texttt{SATzilla'#1-like}}
\newcommand{\isaclike}{\texttt{ISAC-like}}
\newcommand{\ssslike}{\texttt{3S-like}}
\newcommand{\measplike}{\texttt{ME-ASP-like}}

\newcommand{\sbs}{\texttt{Single Best}}

\newcommand{\gga}{\texttt{GGA}}
\newcommand{\paramils}{\texttt{ParamILS}}
\newcommand{\smac}{\texttt{SMAC}}

\newcommand{\comp}{\texttt{Comp-13-Set}}
\newcommand{\foliosuite}{\texttt{Potassco-Set}}

\newcommand{\lpsat}{\texttt{lp$2$sat}}
\newcommand{\lpbv}{\texttt{lp$2$bv}}
\newcommand{\lpmip}{\texttt{lp$2$mip}}
\newcommand{\wasp}{\texttt{wasp}}
\newcommand{\cmodels}{\texttt{cmodels}}
\newcommand{\smodels}{\texttt{smodels}}
\newcommand{\idp}{\texttt{IDP3}}

\newcommand{\expertP}{\texttt{Expert-portfolio}}
\newcommand{\sotaP}{\texttt{SOTA-portfolio}}
\newcommand{\hydraP}{\texttt{Hydra-like-portfolio}}
\newcommand{\isacP}{\texttt{ISAC-like-portfolio}}

\newcommand{\lmc}[1]{\textcolor{red}{#1}}
\renewcommand{\lmc}[1]{#1}

\newcommand{\iclp}[1]{\textcolor{red}{#1}}
\renewcommand{\iclp}[1]{#1}

  \title[Advances in Algorithm Selection for ASP]
        {\claspfolioTwo\ : Advances in Algorithm Selection for Answer Set Programming}

  \author[H. Hoos \and M. Lindauer \and T. Schaub]{%
    Holger Hoos$^1$ and Marius Lindauer$^{2,3}$ and Torsten Schaub$^3$
    \\
    $^1$University of British Columbia, Canada \quad $^2$University of Freiburg, Germany \quad $^3$University of Potsdam, Germany
		}

\submitted{[n/a]}
\revised{[n/a]}
\accepted{[n/a]}

\begin{document}

\maketitle
\begin{abstract}
Building on the award-winning, portfolio-based ASP solver \claspfolio{}, 
we present \claspfolioTwo{}, a modular and open
solver architecture that integrates several different portfolio-based
algorithm selection approaches and techniques.  
The \claspfolioTwo{} solver framework supports various 
feature generators, solver selection approaches, solver portfolios, as well
as solver-schedule-based pre-solving techniques.  The default configuration of
\claspfolioTwo{} relies on a light-weight version of the ASP solver \clasp{}
to generate static and dynamic instance features.  The
flexible open design of \claspfolioTwo{} is a distinguishing factor even beyond
ASP.  As such, it provides a unique framework for comparing and combining
existing portfolio-based algorithm selection approaches and techniques 
in a single, unified framework. Taking advantage of this, 
we conducted an extensive experimental study to assess the impact
of different feature sets, selection approaches and base solver portfolios.
In addition to gaining substantial insights into the utility of the various approaches and techniques,
we identified a default configuration of \claspfolioTwo{} that achieves substantial performance gains not
only over \clasp{}'s default configuration and the earlier version
of \claspfolio{}, but also over manually tuned configurations of \clasp{}.
\end{abstract}

\section{Introduction}
\label{cf:introduction}

Answer Set Programming (ASP;~\cite{baral02a}) has become a popular approach to declarative problem solving.
This is mainly due its combination of a rich and simple modeling language with high performance solving technology.
ASP decouples problem specifications from solving algorithms;
however, 
modern ASP solvers are known to be sensitive to search configurations
-- a phenomenon that is common to advanced Boolean constraint processing techniques.
To avoid the necessity of manual solver configuration, a substantial amount of research was thus devoted to 
automated algorithm configuration and selection approaches,
as we detail in Section~\ref{cf:relatedwork};
in ASP, we find works by~\citeN{gekakascsczi11a}, \citeN{mapuri12a}, \citeN{silisc12a}, \citeN{mapuri13a} and \iclp{\citeN{hokalisc14a}},
 and in particular the two portfolio-based systems
\claspfolio{}~\cite{gekakascsczi11a} and \measp~\cite{mapuri13a}.
The idea of such portfolio-based systems is to train classifiers on features of benchmark instances in order to predict the putatively best solver
from a given solver portfolio.
The portfolio of solvers used in this approach may consist of distinct configurations of the same solver or contain different solvers.

In what follows,
we describe the new portfolio-based ASP system \claspfolio{}, whose earlier
version $1.0$ won first, second, and third places at various ASP competitions.
Version $0.8$ of \claspfolio{} was briefly described in a short paper by \citeN{gekakascsczi11a}
\iclp{and is conceptually identical to the first stable release of version $1.0$.}
The key design features of this prototype were 
(i) feature generation using a light-weight version of the ASP solver \clasp, the original \claspre\ system,
(ii) performance estimation of portfolio solvers via support vector regression, and
(iii) a portfolio consisting of different \clasp~configurations only.
In contrast to this rigid original design,
the new version 2 of \claspfolio{}
provides a modular and open architecture (Section~\ref{cf:framework})
that allows for integrating several different approaches and techniques.
This includes
(i) different feature generators,
(ii) different approaches to solver selection,
(iii) variable solver portfolios, as well as
(iv) solver-schedule-based pre-solving techniques.
The default setting of \claspfolioTwo{} relies on an advanced version of \claspre{} (Section~\ref{cf:features}),
a light-weight version of \clasp{} that produces statistics based on which numerous static and dynamic instance features are generated.

The flexible and open design of \claspfolioTwo{} is a distinguishing factor even beyond ASP.
As such, it provides a unique framework for comparing and combining existing approaches and techniques in a uniform setting.
We take advantage of this and conduct an extensive experimental study comparing the influence of different options regarding (i), (ii), and (iii).
In addition to gaining insights into the impact of the various approaches and techniques,
we identify distinguished options showing substantial performance gains not only over \clasp{}'s default configuration 
but moreover over manually tuned configurations of \clasp.
\iclp{\claspfolioTwo{} is  $19$-$51\%$ faster than the best known static \clasp{} configuration
and also $14$-$37\%$ faster than \claspfolio{} $1.0$,}
as shown in Table~\ref{cf:final} at the end of the paper.
To facilitate reproducibility of our results and to promote the use
of high-performance ASP solving technology, 
we have made \claspfolioTwo{} publicly available as open-source software at \url{http://potassco.sourceforge.net/#claspfolio}.


\section{Related Work}
\label{cf:relatedwork}

Our work continues a long line of research that can be traced back to John Rice's seminal 
work on algorithm selection~\cite{rice76a} on one side, and to work by \citeN{huluho97a} 
on parallel algorithm portfolios on the other side.
Especially on SAT problems,
automatic algorithm selectors have achieved impressive performance improvements in the last decade.
\satzilla{}~ (\citeANP{xuhuhole08a} \citeyearNP{xuhuhole08a,xuhole07a,xuhuhole09a,xuhuhole11a,xuhuhole11b})
predicted algorithm performance by means of ridge regression until $2009$
and nowadays uses a pairwise voting scheme based on random forests;
\isac{}~\cite{kamaseti10a} clusters instances in the instance feature space
and uses a nearest neighbour approach on cluster centers for algorithm selection;
\sss{}~\cite{kamasasase11a,masasase13a} uses $k$-NN in the feature space 
and introduces pre-solving schedules computed by Integer Linear Programming 
and cost-sensitive clustering;
\snapp{}~\cite{comamesu13a} predicts algorithm performance based on instance features
and chooses an algorithm based on the similarity of the predicted performances.
\iclp{All these systems are specialized on a single approach. 
They are highly efficient but do not provide a uniform setting,
that is, different inputs and different performance metrics.}

\iclp{Apart from SAT, there exist several algorithm selectors for other problems.} 
Following the original \claspfolio{} of \citeN{gekakascsczi11a} approach,
\citeN{mapuri12a} presented \measp, 
a multi-engine algorithm selector for ASP with an instance feature generator for syntactic features.
Similarly, \aqme{}~\cite{pultac07a} is a multi-engine selector for QSAT.
\cphydra{}~\cite{mahehonusu08a} selects a set of CSP solvers based on case-based reasoning
and schedules them heuristically. 
\stonesoup{}~\cite{sebrgahe12a,heroka11a} uses greedy hill climbing to find algorithm schedules for planning problems.
\aspeed{}~\cite{hokalisc14a} also computes algorithm schedules,
but takes advantage of the modeling and solving capabilities of ASP to find timeout-minimal schedules.


Related to our work on a more general level, 
\citeN{huxuhole12a} gave an overview over runtime prediction techniques,
which is also used in some algorithm selection approaches, e.g., \satzilla{09}.
A comparison of different machine learning algorithms for algorithm selection 
was presented by \citeN{kogemi12a}.
Based on these results,
\citeN{kotthoff13a} introduced \llama{}, Leveraging Learning to Automatically Manage Algorithms,
a \iclp{flexible} framework that provides functionality to train and assess the performance of \iclp{different} algorithm selection techniques.

\section{Generalized Algorithm Selection Framework}
\label{cf:framework}


\claspfolioTwo{}'s new algorithm framework combines the flexibility of \llama{} with
additional state-of-the-art techniques and produces an executable algorithm selection solver.
As such, it provides a unique framework for comparing and combining existing approaches and techniques in a uniform setting. 
Furthermore, the new design of \claspfolioTwo{} follows the idea of Level $4$ of \textit{programming by optimisation}~\cite{hoos12a}:
{\em ``The software-development process is centered on the idea of providing design choices 
and alternatives in all parts of a project that might benefit from them; 
design choices that cannot be justified convincingly are not made prematurely.''}

A further distinguishing feature of the \claspfolioTwo{} framework is the efficient and deep integration of an algorithm scheduling system, 
viz.~\aspeed{}~\cite{hokalisc14a},
into an algorithm selection framework \iclp{to compute a static pre-solving schedule}. 
\claspfolioTwo{} uses \aspeed{} to determine the running times used within pre-solving schedules.
Thereby, it considers the estimated quality of the algorithm \iclp{selector} to 
\iclp{determine the running time of the complete pre-solving schedule.}  
\iclp{This also allows us to integrate the pre-solving strategies of \satzilla{} and \sss{}.} 

\begin{figure}[tp]
\scalebox{0.85}{
\begin{tikzpicture}[node distance=0.6cm]
	\node (Instance) [data] {Training Instances};
	\node (Solvers) [data, right=of Instance] {Algorithms};
	\node (Runtimes) [activity, below=of Solvers, text width = 9em, yshift=-1.3em] {Assess Performance};
	\node (Features) [activity, fill=red!20, below=of Instance, text width = 9em, yshift=-1.3em] {Compute Features};
	\node (claspre) [data, left=of Features] {\claspre};

	\draw[myarrow] (Instance) -- (Runtimes.north west);
	\draw[myarrow] (Instance) -- (Features.north);
	\draw[myarrow] (Solvers) -- (Runtimes.north);
	\draw[myarrow] (claspre) -- (Features);
	
	\node (FeatPre) [activity, fill=red!20, below=of Features, text width = 9em, yshift=-1.3em] {Feature Preprocessing};
	\node (TimePre) [activity, fill=red!20, below=of Runtimes, text width = 9em, yshift=-1.3em] {Performance Preprocessing};
	\node (Training) [activity, fill=red!20, below=of TimePre, text width = 9em, yshift=-0.4em] {Train Scoring Model};

	\node (CVML) [activity, right=of TimePre, fill=yellow!20, text width = 11em, xshift=1em] {Performance Estimation};
	\node (Aspeed) [activity, below=of CVML, fill=yellow!20, text width = 9em, yshift=-0.5em] {Pre-Solving Schedule\\ by \aspeed};

	
	\draw[myarrow, bend left=20] (Runtimes) |- ++(6.25,0.0) |- (Aspeed);
	\draw[myarrow] (Runtimes) -- (TimePre);
	\draw[myarrow] (Features) -- (FeatPre);
	\draw[myarrow] (TimePre) -- (Training);
	\draw[myarrow] (FeatPre) |- (Training);
	\draw[myarrow] (CVML) -- (Aspeed);
	
	\node (RunSBS) [activity, below=of Aspeed, text width=10em, fill=yellow!20, yshift=-3.0em] {Run Pre-Solving\\ Schedule};
	\node (Run) [activity, fill=red!20, below=of RunSBS] {Run Best Scored Algorithm};
	
	\draw[myarrow, dashed, shorten <=0.23cm] (TimePre) -- node [above] {I} (CVML);
	\draw[myarrow] (Aspeed) -- (RunSBS);
	\draw[myarrow] (RunSBS) -- node [right] {if not successful} (Run);
	
	\node (Predict) [activity, fill=red!20, left=of RunSBS, xshift=0.0em] {Score Algorithms};
	\node (TestFeatures) [activity, fill=red!20, left=of Predict] {Compute Features};
	\node (TestInstance) [data, left=of TestFeatures] {(New) Instance};
	\node (Backup) [activity, fill=red!20, below=of TestFeatures] {Run Backup Algorithm};
	
	\draw[myarrow] (Training.south) -- ($(Predict.north)+(-0.62,0.0)$);
	\draw[myarrow] (TestInstance) -- (TestFeatures);
	\draw[myarrow] (TestFeatures) -- (Predict.west);
	\draw[myarrow] (Predict) -- (RunSBS.west);
	\draw[myarrow] (TestFeatures) -- node [left] {failed} (Backup);		
	\draw[myarrow] (claspre.south) -- ($(TestInstance.north)+(-0.8,0.2)$) -| (TestFeatures.north);	
	
	\draw[myarrow, dashed] (Aspeed) -- node [above, pos=0.63] {II} (Training);
	
    \begin{pgfonlayer}{background}
    
    	\path (Instance -| Instance.west)+(-0.25,0.55) node (resUL) {};
    	\path (Solvers.east |- Solvers.south)+(0.75,-0.5) node(resBR) {};
    	\path [rounded corners, draw=black!50, dashed] (resUL) rectangle (resBR);
		\path (Solvers.east |- Solvers.south)+(-0.0,-0.3) node [text=black!60] {Resources};
    
    	\path (Features -| Features.west)+(-0.25,0.55) node (dataUL) {};
    	\path (Runtimes.east |- Runtimes.south)+(0.75,-0.5) node(dataBR) {};
    	\path [rounded corners, draw=black!50, dashed] (dataUL) rectangle (dataBR);
		\path (Runtimes.east |- Runtimes.south)+(-0.3,-0.3) node [text=black!60] {Data Collection};
    	
    	\path (TimePre -| FeatPre.west)+(-0.25,0.55) node (trainUL) {};
    	\path (Training.east |- Aspeed.south)+(0.25,-0.5) node(trainBR) {};
    	\path [rounded corners, draw=black!50, dashed] (trainUL) rectangle (trainBR);
    	\path (Training.east |- Aspeed.south)+(-0.5,-0.3) node [text=black!60] {Prediction};
    	
    	\path (CVML -| CVML.west)+(-0.25,0.55) node (trainUL) {};
    	\path (CVML.east |- Aspeed.south)+(0.25,-0.5) node(trainBR) {};
    	\path [rounded corners, draw=black!50, dashed] (trainUL) rectangle (trainBR);
    	\path (Aspeed.east |- Aspeed.south)+(-0.2,-0.3) node [text=black!60] {Scheduling};
    	
    	\path (CVML -| FeatPre.west)+(-0.5,0.75) node (trainUL) {};
    	\path (CVML.east |- Aspeed.south)+(1.0,-1.1) node(trainBR) {};
    	\path [rounded corners, draw=black!50, dashed] (trainUL) rectangle (trainBR);
    	\path (Aspeed.east |- Aspeed.south)+(0.7,-0.9) node [text=black!60] {Training};
    	
    	\path (TestFeatures.west)+(-0.2,0.55) node (testUL) {};
    	\path (Run.east |- Run.south)+(1.5,-0.5) node(testBR) {};
    	\path [rounded corners, draw=black!50, dashed] (testUL) rectangle (testBR);
    	\path (Run.east |- Run.south)+(0.9,-0.3) node [text=black!60] {Solving};
    	
    \end{pgfonlayer}
\end{tikzpicture}
}
\caption{General workflow of \claspfolioTwo{}. 
		Objects such as algorithms and instances are shown as rectangles, and
		activities are depicted as rectangles with rounded corners.
		Activities related to algorithm are tinted red
		and activities related to algorithm schedules yellow.   
} 
\label{dia:cf:framework}
\end{figure}
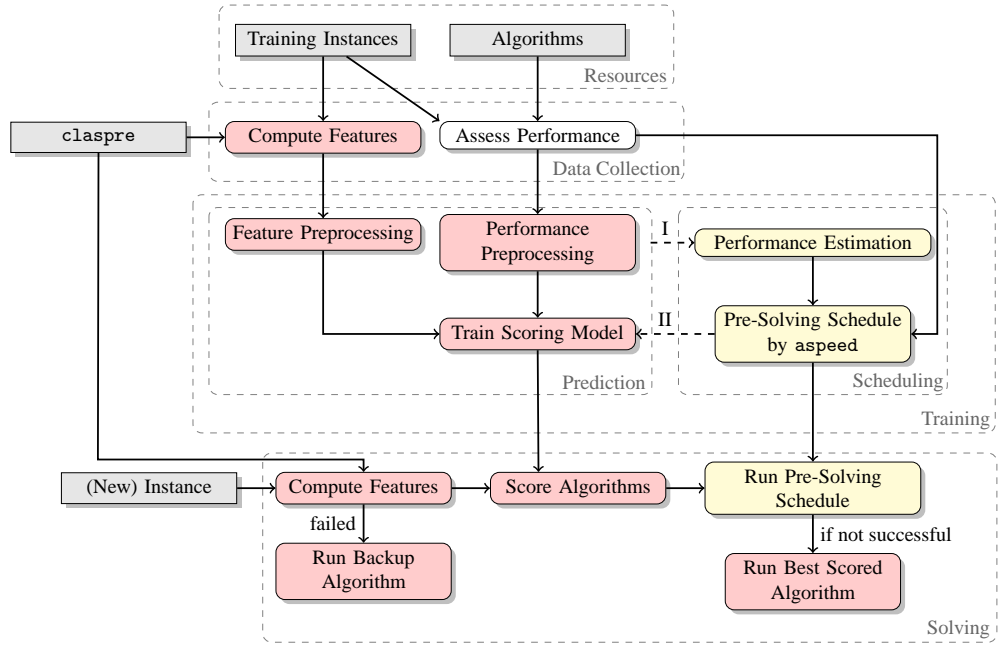

The general workflow underlying \claspfolioTwo{} consists of collecting training data,
learning a prediction model and training a pre-solving schedule;
the portfolio-based ASP solver thus obtained
solves a given problem instance with the pre-solving schedule and a solver selected by the prediction model.
In what follows,
we describe how this workflow is implemented efficiently in \claspfolioTwo{}; see Figure \ref{dia:cf:framework}.

\vspace*{-2mm}

\paragraph{1. Resources.} 
To train an algorithm selector,
\emph{training instances} and a portfolio of \emph{algorithms} are required.
Algorithm selection is based on the assumption 
that the given \emph{training instances} are representative for 
the instances to be solved using the trained algorithm selection solver. 
In addition, a portfolio, i.e., a set of \emph{algorithms} with complementary strengths
(e.g., high-performance solvers used in a competition), 
provides the basis for algorithm selectors to efficiently solve a large variety of instances.

\vspace*{-2mm}

\paragraph{2. Data Collection.} 
An algorithm selection task is defined 
based on the performance of all algorithms on all training instances (\emph{Assess Performance}), 
instance features for each instance (\emph{Compute Features})
and the costs for feature computation define an algorithm selection task.
\claspfolioTwo{} supports several feature generators, 
of which \claspre{} is used by default.

\vspace*{-2mm}

\paragraph{3. Training.} The training phase of \claspfolioTwo{} makes use of two distinct components: 
\emph{Prediction} 
and \emph{Scheduling}. 
Both components can also be used separately in \claspfolioTwo{}.  
  		
The \emph{Prediction} component of \claspfolioTwo{} involves 
\emph{feature pre-processing}, e.g., feature normalization and feature selection, 
and \emph{performance pre-processing}, 
e.g., performance score transformation
and algorithm filtering\footnote{%
Algorithm filtering removes components of the portfolio given some strategy,
e.g., algorithms with a marginal contribution on virtual best solver performance of $0$ can be removed.
In \cite{xuhuhole08a}, this is called solver subset selection
and in \cite{mapuri12a}, solver selection.
}.
Based on the preprocessed data, a \emph{scoring model} is learned, 
which maps the feature vector for a given problem instance to scores for all \emph{algorithms}
such that algorithms expected to perform well on the given instances are assigned better scores. 

The \emph{Scheduling} component of \claspfolioTwo{} computes a timeout-minimal pre-solving schedule 
using \aspeed{}~\cite{hokalisc14a},
where each algorithm gets a (potentially zero) time slice of the overall runtime budget available for solving a
given problem instance.
If the \emph{prediction} component is not used,
the schedule consists only of the given \emph{algorithms}.
If the \emph{prediction} component is used,
cross validation is used to obtain an unbiased estimate of the performance (\emph{Performance Estimation}) 
of the \emph{prediction} component (Arrow I).
The resulting performance estimate of the prediction component is used as an additional simulated algorithm in the schedule generation process. 
All components of the schedule except the simulated one form the pre-solving schedule used in \claspfolioTwo{}.
If the \emph{prediction} performs well, the pre-solving schedule may be empty 
\iclp{because the pre-solving schedule cannot perform better than a perfect predictor, i.e., the selection of the best solver.}
In contrast, if \emph{prediction} performs very poorly (e.g., as a result of non-informative instance features),
the simulated algorithm may be assigned a time slice of zero seconds and the prediction component is de facto ignored in the \emph{solving} step. 

Like \satzilla{}~\cite{xuhuhole08a}, \claspfolioTwo{} allows to ignore instances solved by the pre-solving schedule (Arrow II)
when learning the scoring model,
such that the resulting model is focused on the harder instances not solved by the pre-solvers
that are actually subject to algorithm selecting during the solving phase.

\vspace*{-2mm}

\paragraph{4. Solving} a \emph{(new) instance} starts with the computation of its features. 
If feature computation fails, e.g., because it requires too much time, 
a \emph{backup solver} is used to solve the instance.
Otherwise, the scoring model is used to \emph{score} each algorithm of the portfolio based on the computed feature vector.
If the algorithm with the best score is part of the pre-solving schedule, it is removed from the schedule,
because running the same algorithm twice does not increase the solving probability (when using deterministic algorithms
like \clasp{}). 
Next, the \emph{pre-solving schedule} is executed.\footnote{
Unlike this, \satzilla{} runs the pre-solving schedule first and then computes the instance features,
because the feature computation can be costly in SAT and the pre-solving schedule can solve the instance without incurring this cost.
However, this does not permit removal of the selected solver from the pre-solving schedule.
}
If at the end of executing the pre-solving schedule, the instance  has not been solved, 
the algorithm with the highest score is run for the remainder of the overall time budget.


\section{\claspre{}: Instance Features for ASP}
\label{cf:features}

\begin{table}[tb]
\begin{framed}
\vspace*{-2mm}
\begin{multicols}{3}
\noindent
Tight\\
$\#$ Problem Variables\\
$\#$ Free problem Variables\\
$\#$ Assigned problem Variable\\
$\#$ Constraints\\
$\#$ Constraints / $\#$Variables\\
$\#$ Created Bodies\\
$\#$ Program Atoms\\
$\#$ SCCs\\
$\#$ Nodes in positive BADG\\
$\#$ Rules\\
$\#,\%$ Normal Rules\\
$\#,\%$ Cardinality Rules\\
$\#,\%$ Choice Rules\\
$\#,\%$ Weight Rules\\
$\%$ Negative body Rules\\
$\%$ Positive body Rules\\
$\%$ Unary Rules\\
$\%$ Binary Rules\\
$\%$ Ternary Rules\\
$\%$ Integrity Constraints\\
$\#$ Equivalences\\
$\#,\%$ Atom-Atom Equivalences\\
$\#,\%$ Body-Body Equivalences\\
$\#,\%$ Other Equivalences\\
$\#,\%$ Binary Constraints\\
$\#,\%$ Ternary Constraints \\
$\#,\%$ Other Constraints 
\end{multicols}
\vspace*{-4mm}
\end{framed}
\caption{38 static features computed by \claspre{} ($\#$ = number, $\%$ = fraction, SCCs = Strongly Connected Components, BADG = Body-Atom-Dependency Graph)}
\label{tab:features:static}
\end{table}

\begin{table}[tb]
\begin{framed}
\vspace*{-2mm}
\begin{multicols}{3}
\noindent
$\#$ Choices\\
$\#$ Conflicts / $\#$Choices\\
$\varnothing$ conflict level\\
$\varnothing$ LBD level\\
$\#,\%$ Learnt conflict nogoods\\
$\#,\%$ Learnt loop nogoods\\
$\#,\%$ Literals conflict nogoods\\
$\#,\%$ Literals loop nogoods\\
$\#,\%$ Removed nogoods\\
$\#,\%$ Learnt binary nogoods\\
$\#,\%$ Learnt ternary nogoods\\
$\#,\%$ Learnt other nogoods\\
Longest backjump (bj)\\
$\#,\varnothing$ Skipped levels while bj\\
running average Conflict level\\
running average LBD level
\end{multicols}
\vspace*{-4mm}
\end{framed}
\caption{25 dynamic features computed (at each restart) by \claspre{} \newline  ($\#$ = number, $\%$ = fraction, $\varnothing$ = average, LBD = Literal Blocking Distance)}
\label{tab:features:dynamic}
\end{table}

The entire concept of algorithm selection is based on instance features
which characterize benchmark instances 
and allow for predicting the putatively best solver from a given portfolio.
These instance features should be \iclp{cheap-to-compute} to save as much time as possible for the actual solving process,
but should also provide sufficient information to distinguish 
between (classes of) instances for which different solvers or solver configurations work best.

For feature generation, \claspfolioTwo{} uses \claspre{} in its default configuration.
\claspre{} is a light-weight version of \clasp~\cite{gekakaosscsc11a} 
that extracts instance features of ground ASP instances in \smodels~format~\cite{lparseManual},
using \clasp{}'s internal statistics.
The features determined by \claspre{} can be grouped into static and dynamic ones.
The former are listed in Table~\ref{tab:features:static} and include $38$ properties, such as number of constraints.
Beyond that, \claspre{} performs a limited amount of search 
to collect dynamic information about solving characteristics.
These dynamic features are computed after each restart of the search process,
where restarts are performed after a fixed number of conflicts.
Thereby, $25$ dynamic features (Table \ref{tab:features:dynamic}) are extracted after each restart, such as 
the average number of conflict levels skipped while back-jumping.

The number of restarts performed is a parameter of \claspre{}.
More restarts lead to longer feature vectors that may contain more information.
The number of restarts and number of conflicts between restarts determine
the time used by \claspre{} for feature computation
We note that the pre-processing and search performed by \claspre{} can actually 
solve a given ASP instance.
The probability of this happening increases with the length of the search 
performed within \claspre{};
however, at the same time, long runs of \claspre{} reduce
the time available for running solvers from the portfolio.


\section{Empirical Performance Analysis}
\label{cf:benchmarks}

As previously described,
\claspfolioTwo{}'s modular and open architecture (Section~\ref{cf:framework})
allows for integrating several different approaches and techniques,
including
(i) different feature generators,
(ii) different approaches to solver selection, as well as
(iii) variable solver portfolios.
Taking advantages of this flexibility, we
conducted an extensive experimental study to assess the efficacy of the various choices 
on large and representative sets of ASP instances.

Training data of \claspfolioTwo{} is stored in the algorithm selection data format 
developed by the \emph{COSEAL Group},\footnote{\url{https://code.google.com/p/coseal}}
an international group of experts in the field of algorithm selection and configuration.
Detailed experimental results and the source code of \claspfolioTwo{} are available at \url{http://www.cs.uni-potsdam.de/claspfolio}.
Our empirical analysis makes use of commonly used techniques from statistics and machine learning
(see, e.g., \cite{bishop07a}).


\subsection{Setup}

All our experiments were performed on a computer cluster
with dual Intel Xeon E$5520$ quad-core processors ($2.26$~GHz, 8192 KB cache) and $48$~GB RAM per node,
running Scientific Linux (2.6.18-308.4.1.el5).
Each algorithm run was limited to a runtime cutoff of $600$ CPU seconds \iclp{and to a memory cutoff of $6$~GB}.
Furthermore,
we used permutation tests with $100\ 000$ permutations and significance level $\alpha=0.05$ 
to our performance metrics, the ($0$/$1$) timeout scores, the PAR$10$ scores and the PAR$1$ scores,\footnote{ %
PAR$X$ is the penalized average runtime penalizing timeouts by $X$ times the runtime cutoff.}
to asses the statistical significance of observed performance differences.


\subsection{Instance Sets}

We used all instances submitted to the $2013$ ASP Competition in the \texttt{NP}
category that could be grounded with \gringo{} ($3.0.5$) within $600$ CPU seconds and $6$ GB memory.
The resulting instance set consists of $2214$ instances from $17$ problem classes;
we call it \comp{}.
As an even more heterogeneous instance set, 
we used the ASP \foliosuite{} introduced by \citeN{hokascsc13a};
it consists of $2589$ instances from $105$ problem classes
and includes instances from the ASP competitions organized in
$2007$ (SLparse track), 
$2009$ (with the encodings of the Potassco group) and
$2011$ (decision NP-problems from the system track), as well as 
several instances from the ASP benchmark collection platform \textit{asparagus}.\footnote{\url{http://asparagus.cs.uni-potsdam.de}}
All instances were grounded with \gringo{}, and the grounding time was not counted towards solving the instances.

Each instance set was randomly split into equally sized, disjoint training and test set;
only the training sets were used in the process of building algorithm portfolios.
The resulting \claspfolioTwo{} solvers were evaluated on the hold-out test sets.
\iclp{We also used the training instances to determine the best \claspfolioTwo{} configuration (Subsection 5.3).
To assess the performance of \claspfolioTwo{} (Subsection 5.6), we used a $10$-fold cross validation on the test set.
Notice that we cannot use the training set for \claspfolioTwo{} to obtain an unbiased learned model,
because the algorithm portfolios have an optimistic performance estimation on the training set on which they were build.}

\subsection{Building Algorithm Portfolios}
\label{sec:sub:cf:portfolio}

In addition to a set of training instances,
a portfolio (i.e., a set) of algorithms is required to construct a portfolio solver.
\claspfolioTwo{} can handle portfolios containing different solvers
as well as different configurations of a given solver, all of which 
are viewed as individual ASP solvers.
We investigated the following portfolios of ASP solvers:
\begin{itemize}
  \item \expertP{} of four \clasp{} (2.1.3) configurations designed by Benjamin Kaufmann (configurations: 
    \textit{frumpy} (default), \textit{jumpy}, \textit{handy} and \textit{crafty})
  \item \sotaP{}~\cite{mapuri12a}: 
    non-portfolio solvers participating in the $2013$ ASP Competition%
    \footnote{\idp{} was removed from the portfolio because it was strongly dominated by all other solvers.} 
    and in addition, the well-established solvers \cmodels{} and \smodels{}; in detail:
    \clasp{}~\cite{gekakaosscsc11a},
    \cmodels{}~\cite{gilima06a},
    \lpbv{}~\cite{ngjani11a},
	\lpmip{}~\cite{lijani12a},
	\lpsat{}~\cite{janhunen06a},
    \smodels{}~\cite{siniso02a},
    and \wasp{}~\cite{aldofaleri13a}
  \item \hydraP{}~\cite{xuhole10a,xuhuhole11a} of \clasp{} (2.1.3) configurations
  \item \isacP{}~\cite{kamaseti10a} of \clasp{} (2.1.3) configurations
\end{itemize}

\expertP{} and \sotaP{} are portfolios manually constructed by experts.
In contrast, \hydra{} and \isac{} are automatic methods for constructing portfolios
using algorithm configurators, e.g., \paramils{}~\cite{huhost07a}, \gga{}~\cite{anseti09a} or \smac{}~\cite{huhole11b}.
They generate a portfolio of configurations of a given solver
by determining configurations that complement each other well on a given set of training instances,
with the goal of optimizing the performance of the portfolio under the idealized assumption
of perfect selection; this performance is also called 
the virtual best solver (vbs) or oracle performance of the portfolio.

An implementation of \hydra{} that can be applied to solvers for arbitrary problems has not yet been published by \citeANP{xuhole10a};
therefore, we have implemented our own version of \hydra{} (in consultation with the authors),
which we refer to as \hydraP{} in the following.
Also, since the only published version of \isac{} ($2.0$) does not include algorithm configuration,
we reimplemented the part of \isac{} responsible for portfolio generation,
dubbed \isacP{}.
In contrast to the original \isac{}, which performs g-means clustering,
\isacP{} uses k-means clustering, 
where the number of clusters is determined by using cross-validation 
to optimize the scoring function
of the k-means procedure
(following \citeN{hokascsc13a}).

Using this approach, \isacP{} found $15$ clusters for \comp{} and $11$ clusters for \foliosuite{},
inducing $15$ and $11$ configuration tasks, respectively.
To obtain a fair comparison, we allocated the same time budget to \hydraP{}
and allowed it to perform $15$ and $11$ iterations, respectively (each consisting of one configuration task).
The configuration process performed by \smac{} ($2.06.01$; \citeNP{huhole11b}) on each cluster 
and in each \hydra{} iteration, respectively, was allocated $120\ 000$ CPU seconds,
i.e., $200$ times the target algorithm cutoff time, and $10$ independent repetitions,
from which the result with the best PAR10 score on the given training set was selected.
\iclp{\smac{} optimized PAR$10$.}

\begin{table}
\centering
\begin{tabular}{l ccc ccc}
\hline
\hline
			& \multicolumn{3}{c}{\comp{}}						& \multicolumn{3}{c}{\foliosuite{}}\\
			& \#TOs 	& PAR$10$ 	& PAR$1$ 						& \#TOs	& PAR$10$ 	& PAR$1$\\
			\cline{2-4} \cline{5-7}\\
\expertP 	& $360$		& $2169$	& $255$						& $100$		& $491$	& $74$\\
\sotaP 		& $335$		& $1866$	& $231$						& $111$		& $538$	& $75$\\
\hydraP 	& $326$		& $1798$	& $207$						& $\mathbf{82}$	& $\mathbf{400}$	& $\mathbf{58}$\\
\isacP 		& $\mathbf{313}$ & $\mathbf{1724}$	& $\mathbf{196}$& $99$		& $476$	& $63$\\
\hline
\hline
\end{tabular}
\caption{Virtual best solver (VBS) performance of portfolio building approaches on test sets.
		 Results shown in boldface were statistically significantly better than all others within the respective column
		 (according to a permutation test with $100\ 000$ permutations and 
		 $\alpha=0.05$).
}
\label{cf:vbs}
\end{table}

Table~\ref{cf:vbs} shows the performance of the virtual best solvers (i.e., the performance of a perfect algorithm selector) 
for the different considered portfolios.
Interestingly, the results differ qualitatively between two benchmark sets.
While \sotaP{} performs better than \expertP{} on \comp{},
\expertP{} is better on \foliosuite{}.
Furthermore, while for both sets, the automatic generation methods found better performing portfolios than the the manual selected methods,
on the \comp{}, \isacP{} produced a better results than \hydraP{},
and the opposite holds for \foliosuite{}.
Furthermore, unlike conjectured by \citeN{mapuri12a}, a set of configurations of the same, highly parameterized solver (\expertP{}, \isacP{} and \hydraP{}) 
generally did not yield worse performance than a mixed portfolio, such as \sotaP{}.

While we gave \hydra{} the same time budget as \isac{} to find portfolios,
the components added by \hydraP{} in its final three iterations
decreased the number of timeouts only by one on our training and test sets.
Following \citeN{xuhole10a}, \hydra{} would be terminated 
when the performance does not improve on the training set after an iteration.  
Hence, \hydraP{} not only produced a better portfolio on \foliosuite{} than \isac{},
but also does so using less configuration time than 
\isac{}.


\subsection{Feature Sets}

In addition to the \claspre{} feature set presented in Section~\ref{cf:features},
we considered a set of ASP features introduced by \citeN{mapuri13a} 
that is focussed on very efficiently computable syntactic features, such as
number of variables. 
The published version of their feature generator supports only the ASPCore $1.0$~\cite{caiari11a} language
of the  $2011$ ASP Competition.
Our \comp{} consists of instances of the $2013$ ASP Competition in ASPCore $2.0$,
which introduced further language constructs.
Therefore, we re-implemented this feature generator with the help of \citeANP{mapuri13a} 
to be compatible with ASPCore $2.0$.\footnote{
The new feature generator is implement in Python, whereas the original generator was implemented in C++,
which induced an overhead of a factor $2$ in terms of running time on average on ASPCore $1.0$ instances from the $2011$ ASP Competition.
}

One of the most established and investigated feature generators for SAT 
is provided as part of \satzilla{}~\cite{xuhuhole08a}.
ASP instances can be translated to SAT with techniques by \citeN{janhunen06a}, 
using his tool \lpsat{}.
We use a combination of \lpsat{}\footnote{\lpsat{} was used as submitted at the $2013$ ASP Competition.}
with the feature generator of \satzilla{} to generate a set of instance features for ASP instances;
this is the first time, these features are studied in the context of ASP.
Since the full set of \satzilla{} features is very expensive to compute and our SAT encodings can get quite large,
we decided to only use the efficiently computable base features.

\begin{table}
\centering
\footnotesize
\begin{tabular}{l ccccc ccccc}
\hline
\hline
					& \multicolumn{5}{c}{\comp{}}												& \multicolumn{5}{c}{\foliosuite{}}\\
					& Min		& $Q_{0.25}$ 	& Median 		& $Q_{0.75}$ 		& \%TOs 	& Min		& $Q_{0.25}$ 	& Median 		& $Q_{0.75}$ 		& \%TOs\\
					\cline{2-6} \cline{7-11}
\claspresta{}		& $0.04$	& $1.43$	 	& $1.72$		& $8.83$			& $16.2$	& $0.13$		& $0.91$	 		& $1.38$			& $1.72$				& $1.0$\\
\claspredyn{}		& $0.07$	& $1.36$		& $1.72$		& $13.94$			& $16.2$	& $0.18$		& $0.87$	 		& $1.48$			& $1.81$				& $1.1$\\
\measp{}			& $0.04$	& $1.18$		& $1.97$		& $15.97$			& $3.2$		& $0.06$		& $0.83$	 		& $1.10$			& $1.79$				& $0.1$\\		
\lpsat{}			& $0.08$	& $24.88$		& $484.85$		& $600$				& $49.4$	& $0.04$		& $3.81$	 		& $21.82$			& $91.13$				& $14.6$\\
\hline
\hline
\end{tabular}
\caption{Time required for computing the features of a single ASP instance in CPU seconds, with a $600$ seconds runtime cutoff. 
We report minimum (Min), $25\%$ quantile ($Q_{0.25}$), median and $75\%$ quantile ($Q_{0.75}$) of the distribution 
over the respective instance set, as well as the percentage of timeouts (\%TOs).}
\label{cf:feature:overhead}
\end{table}

Table~\ref{cf:feature:overhead} shows 
the runtime statistics for
\claspre{} with static features, \claspresta{},
\claspre{} with static and dynamic features, \claspredyn{}, with $4$ restarts and $32$ conflicts between the restarts,
the (re-implemented) feature generator of \measp{}
and the combination of \lpsat{} and \satzilla's feature generator on 
our full benchmark sets (training + test instances).
\claspresta{} is only slightly faster than \claspre{} with additional dynamic features,
since its search was limited to $128$ conflicts. 
To solve typical ASP instances, searches well beyond $100\,000$ conflicts are often required;
nevertheless, \claspresta{} solved $51$ instances through pre-processing,
and \claspredyn{} solved $123$ instances on \comp{},
$9$ and $400$ instances on \foliosuite{}, respectively. 
The feature generation of \measp{}
was faster, but (unsurprisingly, considering the nature of these features) did not solve any instance.
Because of the substantial overhead of generating translations from ASP to SAT,
the combination of \lpsat{} and \satzilla's feature generator turned out to be substantially slower than the other approaches
and failed to compute the feature vectors of $1094$ instances on \comp{} and $377$ instances on \foliosuite{} within the given
cutoff time.


\subsection{Algorithm Selection Approaches}

As previously mentioned,
\claspfolioTwo{} was explicitly designed to easily integrate several state-of-the-art algorithm selection approaches.
This not only permits us to optimize the performance of \claspfolioTwo{}, but also 
to compare the considered algorithm selection approaches within a controlled environment.
Although our re-implementations may not 
reproduce the original implementations in all details (something that would be difficult to achieve, considering
that sources are not available for some published approaches),
they provide the only freely available, open-source implementations of some of these systems
and thus provide a basis for further analysis and improvements.%
\footnote{As with \hydra{} and \isac{} above, published and trainable, general-purpose implementations of \sss{} and \measp{} 
are not available.}

\begin{table}
\centering
\begin{tabular}{l p{9em}ccc}
\hline
\hline
					& Approach 					& Feat. Norm. 		& Pre-Solver 	& Pre-Solver Time \lmc{[sec]}\\
					\cline{2-5}
\aspeed{}			& static schedule			& none							& $\leq \infty$			& $\leq \infty$\\
\claspfoliolike{1.0}& support vector\newline regression	& $z$-score				& $0$				& $0$\\
\measplike{}		& nearest neighbor			& none							& $0$				& $0$\\
\isaclike{}			& $k$-means clustering		& linear						& $0$				& $0$\\
\ssslike{}			& $k$-NN					& linear						& $\leq \infty$			& $\leq \text{cutoff}/10$\\
\satzillalike{09}{}	& ridge regression			& $z$-score						& $\leq 2$				& $\leq 20$\\
\satzillalike{11}	& voting with random forest	& $z$-score						& $\leq 3$				& $\leq 30$\\
\hline
\hline
\end{tabular}
\caption{Algorithm selection mechanism supported by \claspfolioTwo{}.}
\label{cf:as:overview}
\end{table}


Table~\ref{cf:as:overview} gives an overview 
of the approaches available within \claspfolioTwo{}.
These differ with respect to 
(i) the algorithm selection method, 
(ii) the feature normalization technique,
(iii) the maximal number of pre-solvers used
and 
(iv) the maximal running time allocated to the pre-solving schedule.
In all cases, the pre-solving schedules were computed by \aspeed{},
and hyperparameters of the machine learning techniques were set 
using grid search on training data.


\subsection{Results}
\label{sub:sec:cf:results}

\begin{table}
\centering
\begin{tabular}{l cc cc }
\hline\hline
\multicolumn{5}{c}{Impact of feature set}	\\
\hline
				& \multicolumn{2}{c}{\comp{}}							& \multicolumn{2}{c}{\foliosuite{}}\\
 				& $\mu_{PAR10} \pm \sigma_{PAR10}$ 	 & $\min_{PAR10}$ 	& $\mu_{PAR10} \pm \sigma_{PAR10}$ 	 & $\min_{PAR10}$\\
\cline{2-3} \cline{4-5}
\claspresta{} 	& $\mathbf{2116.3} \pm 128.7$    & $1927.0$			& $638.9 \pm 81.1$      & $490.6$\\
\claspredyn{} 	& $2127.6 \pm 122.6$    & $1931.3$					& $\mathbf{630.8} \pm 78.1$      & $\mathbf{480.0}$\\
\measp{} 	 	& $2138.4 \pm 127.7$    & $\mathbf{1919.4}$			& $661.0 \pm 108.8$     & $486.0$\\
\lpsat{} 	 	& $2240.3 \pm 81.3$     & $2056.9$					& $688.3 \pm 45.6$      & $610.3$\\
\hline\hline
%
%
\multicolumn{5}{c}{Impact of portfolio}	\\
\hline
				& \multicolumn{2}{c}{\comp{}}							& \multicolumn{2}{c}{\foliosuite{}}\\
 				& $\mu_{PAR10} \pm \sigma_{PAR10}$ 	 & $\min_{PAR10}$ 	& $\mu_{PAR10} \pm \sigma_{PAR10}$ 	 & $\min_{PAR10}$\\
\cline{2-3} \cline{4-5}
\expertP{} 	 	& $2251.8 \pm 55.0$     & $2165.0$						& $679.1 \pm 47.7$      & $621.6$\\
\sotaP{} 	 	& $2172.4 \pm 60.6$     & $2072.9$						& $691.9 \pm 55.3$      & $614.7$\\
\hydraP{} 	 	& $2141.5 \pm 160.4$    & $1943.7$						& $\mathbf{609.6} \pm 103.5$     & $\mathbf{480.0}$\\
\isacP{} 	 	& $\mathbf{2056.9} \pm 111.3$    & $\mathbf{1919.4}$	& $638.3 \pm 90.9$      & $526.7$\\
\hline\hline
%
%
%
\multicolumn{5}{c}{Impact of selection mechanism}	\\
\hline
				& \multicolumn{2}{c}{\comp{}}							& \multicolumn{2}{c}{\foliosuite{}}\\
 				& $\mu_{PAR10} \pm \sigma_{PAR10}$ 	 & $\min_{PAR10}$ 	& $\mu_{PAR10} \pm \sigma_{PAR10}$ 	 & $\min_{PAR10}$\\
\cline{2-3} \cline{4-5}
\aspeed{} 	 			& $2292.8 \pm 66.1$ 	 & $2222.0$				& $731.2 \pm 40.8$ 	 & $672.6$\\
\claspfoliolike{1.0} 	& $2152.7 \pm 108.0$ 	 & $1978.6$				& $650.3 \pm 58.3$ 	 & $519.3$\\
\measplike{} 	 		& $2245.3 \pm 77.3$ 	 & $2091.8$				& $753.3 \pm 76.7$ 	 & $656.8$\\
\isaclike{} 	 		& $2100.1 \pm 113.5$ 	 & $1939.5$				& $608.4 \pm 65.7$ 	 & $490.6$\\
\ssslike{} 	 			& $2092.0 \pm 109.2$     & $1927.0$				& $596.0 \pm 57.6$   & $489.1$\\
\satzillalike{09} 	 	& $2120.3 \pm 99.4$ 	 & $1932.6$				& $652.7 \pm 48.2$ 	 & $544.0$\\
\satzillalike{11} 	 	& $\mathbf{2086.4} \pm 125.9$ 	 & $\mathbf{1919.4}$	& $\mathbf{591.1} \pm 62.5$ 	 & $\mathbf{480.0}$\\
\hline\hline
\end{tabular}
\caption{Statistics ($\mu$ = average, $\sigma$ = standard deviation, $\min$ = minimum) of PAR10 performance
over all combinations except for the one kept fixed to assess its impact.
\vspace*{-0.4cm}
}
\label{cf:results}
\end{table}


We have assessed the performance of \claspfolioTwo{} on all $112$ combinations of our $4$ feature sets,
$4$ portfolios and $7$ algorithm selection approaches, 
using a cross validation on both test sets.
To study the effect of each design choice,
we collected statistics over the distribution of results by keeping one choice fixed and varying all remaining components;
the results are shown in Table~\ref{cf:results}.
The top part of the table shows results
obtained for using each of the feature sets, 
in terms of average PAR$10$ performance, standard deviation in PAR$10$ performance and best
PAR$10$ performance
over all $28$ combinations of portfolios and selection approaches.
The subsequent parts of Table~\ref{cf:results} show analogous results 
for different portfolios and
selection approaches.

On average, the best feature set was \claspresta{} (the static 
\claspre{} features) on \comp{},
followed by \claspredyn{} (the static + dynamic \claspre{} features),
the feature sets of \measp{} and \lpsat{}.
However, the best \claspfolioTwo{} configuration on \comp{} used \measp{}.
The fact that \claspredyn{} gave worse results than \claspresta{},
although the former is superset of the latter, indicates
that not all features were useful and that feature selection should be used 
to identify a subset of features with highest information content.
On \foliosuite{},
the best average performance and the best performance of any \claspfolioTwo{} configuration
was consistently obtained by using \claspredyn{}.
We believe that the additional dynamic features are necessary 
to distinguish between the larger number of different problem classes in \foliosuite{}.

The results on the impact of the portfolio of algorithms
used as a basis for algorithm selection confirm our assumption 
that the best potential performance, i.e., best VBS performance,
is a good indicator of the actual performance achieved by a high-performance
selection approach.
On \comp{}, \isacP{} achieved the best performance,
while on \foliosuite{}, \hydraP{} yielded even better results.
Furthermore, the portfolios obtained using the two automatic portfolio generation methods, \isac{} and \hydra{},
yielded better results than the manually created ones, \expertP{} and \sotaP{}.
 
As shown in the lower part of Table~\ref{cf:results},
the \satzillalike{11} approach performed best on both benchmark sets,
followed closely by \ssslike{} and \isaclike{}.
\satzillalike{09} and \claspfoliolike{1.0} showed similar, but weaker performance results, 
followed by the \measplike{} approach
and the pure algorithm schedules of \aspeed{}.

\iclp{Overall, the best combination both on the training and test sets of \comp{} was the \measp{} features,
\isacP{} and \satzillalike{11} selection approach,
and \claspredyn{} features, \hydraP{} and \satzillalike{11} selection approach for \foliosuite{}.}

%
%

\begin{table}[t]
\centering
\begin{tabular}{l ccc ccc}
\hline
\hline
							& \multicolumn{3}{c}{\comp{}}						& \multicolumn{3}{c}{\foliosuite{}}\\
							& $\#TOs$ 	& PAR$10$ 	& PAR$1$ 						& $\#TOs$ 	& PAR$10$ 	& PAR$1$\\
							\cline{2-4} \cline{5-7}\\
\clasp{}~(default)			& $577$		& $3168$	& $351$						& $287$		& $1347$	& $176$\\
\clasp{}~(ASP Comp $13$)	& $421$		& $2329$	& $273$						& $150$		& $723$		& $97$\\
\sbs 						& $414$		& $2333$	& $268$						& $150$		& $723$		& $97$\\
\claspfolio{} 1.0	 		& $403$		& $2237$	& $269$						& $134$		& $658$		& $99$\\
\claspfolioTwo{} 			& $\mathbf{353}$		& $\mathbf{1960}$	& $\mathbf{237}$						& $\mathbf{97}$		& $\mathbf{480}$		& $\mathbf{75}$\\
\cline{1-7}
best known VBS				& $313$		& $1724$	& $196$						& $82$		& $400$		& $58$\\
\hline
\hline
\end{tabular}
\caption{Comparison of two \clasp{} configurations,
the \sbs{} solver in all portfolios \iclp{(cf. Subsection~5.3)},
\claspfolio{} 1.0, 
the \claspfolioTwo{} with  \claspredyn{} features, \hydraP{} and \satzillalike{11} approach.
The significantly best performances (except VBS) are shown in boldface
(according to a permutation test with $100\ 000$ permutations and significance level $\alpha=0.05$).
}
\label{cf:final}
\end{table}

\section{Conclusions and Future Work}
\label{cf:conclusion}


Our new, modular \claspfolioTwo{} ASP solver architecture comprises a diverse 
set of portfolio-based algorithm selection techniques, including
feature extractors, manually and automatically constructed base algorithm portfolios, 
algorithm selection mechanisms and solver-schedule-based pre-solving techniques.
As seen from the high-level overview of empirical performance results
in Table~\ref{cf:final}, 
on standard, diverse and heterogeneous sets of ASP benchmarks, 
\claspfolioTwo{} is substantially more robust than the default configuration of \clasp{},
the manual tuned configuration of \clasp{} of the $2013$ ASP Competition,
and than all other assessed individual solvers;
in fact, its performance in terms of PAR10-score lies only about 20\% and 15\% above that of the 
best known oracle on \foliosuite{} and \comp{} benchmark sets, respectively.
The reimplementation of \claspfolio{} $1.0$ in \claspfolioTwo{},
which had a similar performance in preliminary experiments than the original implementation,
achieves also about $14-37\%$ higher PAR10-score than \claspfolioTwo{}.
While the best configuration of \claspfolioTwo{} varies between these two benchmark sets, 
the performance differences are relatively minor: on \comp{}, the best configuration of \claspfolioTwo{} for \foliosuite{}
-- which we also chose as the default configuration for  \claspfolioTwo{} -- achieves
a PAR10-score only about 2.1\% lower than the best configuration for \comp{}, and on \foliosuite{}, 
its PAR10-score is about 9.6\% higher. This configuration  uses
the \claspredyn{} feature set in combination with the \hydraP{} base algorithm portfolio construction
approach and the \satzillalike{11} algorithm selection mechanism, but other feature sets, base algorithm portfolios
and algorithm selection mechanisms also achieve very strong performance.



\section*{Acknowledgments}
T.~Schaub and M.~Lindauer were supported by the DFG projects under SCHA
 550/8-3. H.~Hoos was supported by an NSERC Discovery Grant.

\clearpage
\bibliographystyle{acmtrans}
\bibliography{lit,akku,local,procs}


\end{document}